\documentclass[preprint]{elsarticle}

\usepackage{amssymb}
\usepackage{pifont}
\usepackage{xcolor}
\usepackage{balance}
\usepackage{color}
\usepackage{footnote}
\usepackage{makecell}
\usepackage{rotating}
\usepackage{tabularx}
\usepackage{threeparttable}
\usepackage[ruled,linesnumbered]{algorithm2e}
\usepackage{soul}
\usepackage{bbm}

\usepackage{amsmath,amsfonts}
\usepackage{array}
\usepackage{textcomp}
\usepackage{stfloats}
\usepackage{url}
\usepackage{verbatim}
\usepackage{graphicx}
\usepackage{multirow}
\usepackage{bbding}
\usepackage{gensymb}
\usepackage{booktabs}
\usepackage{subfigure}

\usepackage{color}

\begin{document}

\begin{frontmatter}

\title{Deep Temporal Contrastive Clustering}

\author[1]{Ying Zhong}
\ead{yingzhonghere@hotmail.com}
\author[1]{Dong Huang}
\ead{huangdonghere@gmail.com}
\author[2,3]{Chang-Dong Wang}
\ead{changdongwang@hotmail.com}
\address[1]{College of Mathematics and Informatics, South China Agricultural University, China}
\address[2]{School of Computer Science and Engineering, Sun Yat-sen University, China}
\address[3]{Guangdong Key Laboratory of Information Security Technology, China}

\begin{abstract}
Recently the deep learning has shown its advantage in representation learning and clustering for time series data. Despite the considerable progress, the existing deep time series clustering approaches mostly seek to train the deep neural network by some instance reconstruction based or cluster distribution based objective, which, however, lack the ability to exploit the sample-wise (or augmentation-wise) contrastive information or even the higher-level (e.g., cluster-level) contrastiveness for learning discriminative and clustering-friendly representations. In light of this, this paper presents a \textbf{d}eep \textbf{t}emporal \textbf{c}ontrastive \textbf{c}lustering (DTCC) approach, which for the first time, to our knowledge, incorporates the contrastive learning paradigm into the deep time series clustering research. Specifically, with two parallel views generated from the original time series and their augmentations, we utilize two identical auto-encoders to learn the corresponding representations, and in the meantime perform the cluster distribution learning by incorporating a $k$-means objective. Further, two levels of contrastive learning are simultaneously enforced to capture the instance-level and cluster-level contrastive information, respectively. With the reconstruction loss of the auto-encoder, the cluster distribution loss, and the two levels of contrastive losses jointly optimized, the network architecture is trained in a self-supervised manner and the clustering result can thereby be obtained. Experiments on a variety of time series datasets demonstrate the superiority of our DTCC approach over the state-of-the-art. 
\end{abstract}


\end{frontmatter}


\section{Introduction}

Time series data widely exist in many areas of science and engineering, such as financial data \cite{Cheng2022}, robot sensor data \cite{zhang21TIT}, earthquake data \cite{mich18PRE}, and genomic data \cite{fujita2012functional}. The rapid emergence of time series data has brought new challenges to the current clustering research, where the traditional clustering algorithms are mostly designed for static data and lack the ability to well capture the temporal information \cite{Frey2007,Huang2020,Kang2021,Huang2022,Liang2022}.

In recent years, the deep learning has shown its promising ability in representation learning and clustering for time series data. As one of the earliest deep time series clustering works, Madiraju \cite{madiraju2018deep} proposed a deep temporal clustering (DTC) model, which utilizes an auto-encoder and a temporal clustering layer to learn the clustering assignment. The clustering layer is designed to minimize the Kullback-Leibler (KL) divergence between the predicted distribution and a target distribution \cite{madiraju2018deep}. Further, Ma et al. \cite{ma2019learning} incorporated the $k$-means objective and the temporal reconstruction into a sequence to sequence (seq2seq) model for learning the cluster-specific representations.
The representations learned via the temporal auto-encoder are utilized to form the cluster structure with the guidance of the $k$-means objective. 

 Despite the recent advances, these deep time series clustering methods \cite{madiraju2018deep,ma2019learning} typically learn the representations by considering instance reconstruction or cluster distributions, yet cannot well exploit the sample-wise (or augmentation-wise) information for learning more discriminative representations. More recently, the contrastive learning has attracted significant attention, especially for the image clustering task, which is able to effectively train the deep neural network by leveraging data augmentations and their pair-wise relationships in a self-supervised manner. For example, Li et al. \cite{li2021contrastive} incorporated the contrastive learning into the deep image clustering, and presented the contrastive clustering (CC) method by exploiting the instance-level and cluster-level contrastive learning simultaneously. Li et al. \cite{li2020prototypical} presented an unsupervised representation learning method called prototypical contrastive learning (PCL), which not only obtains the low-level representations for the instance discrimination task, but also encodes the semantic structures acquired by the clustering process into the learned embedding space.  Peng et al. \cite{peng2022crafting} proposed the ContrastiveCrop method, which is able to generate better crops for the Siamese representation learning and further designs a center-suppressed sampling strategy to enlarge the variance of crops. However, these contrastive learning based deep clustering methods \cite{li2021contrastive,li2020prototypical,peng2022crafting} are mostly designed for the image clustering task, which cannot well handle the temporal information and thus are not suitable for time series clustering. Though considerable progress has been made, it is still an open problem how to effectively and jointly incorporate the contrastive learning and the cluster structure learning into the deep time series clustering framework.

 \begin{figure*}[!t]\vskip 0.1 in
 	\begin{center}
 		{\includegraphics[width=1\linewidth]{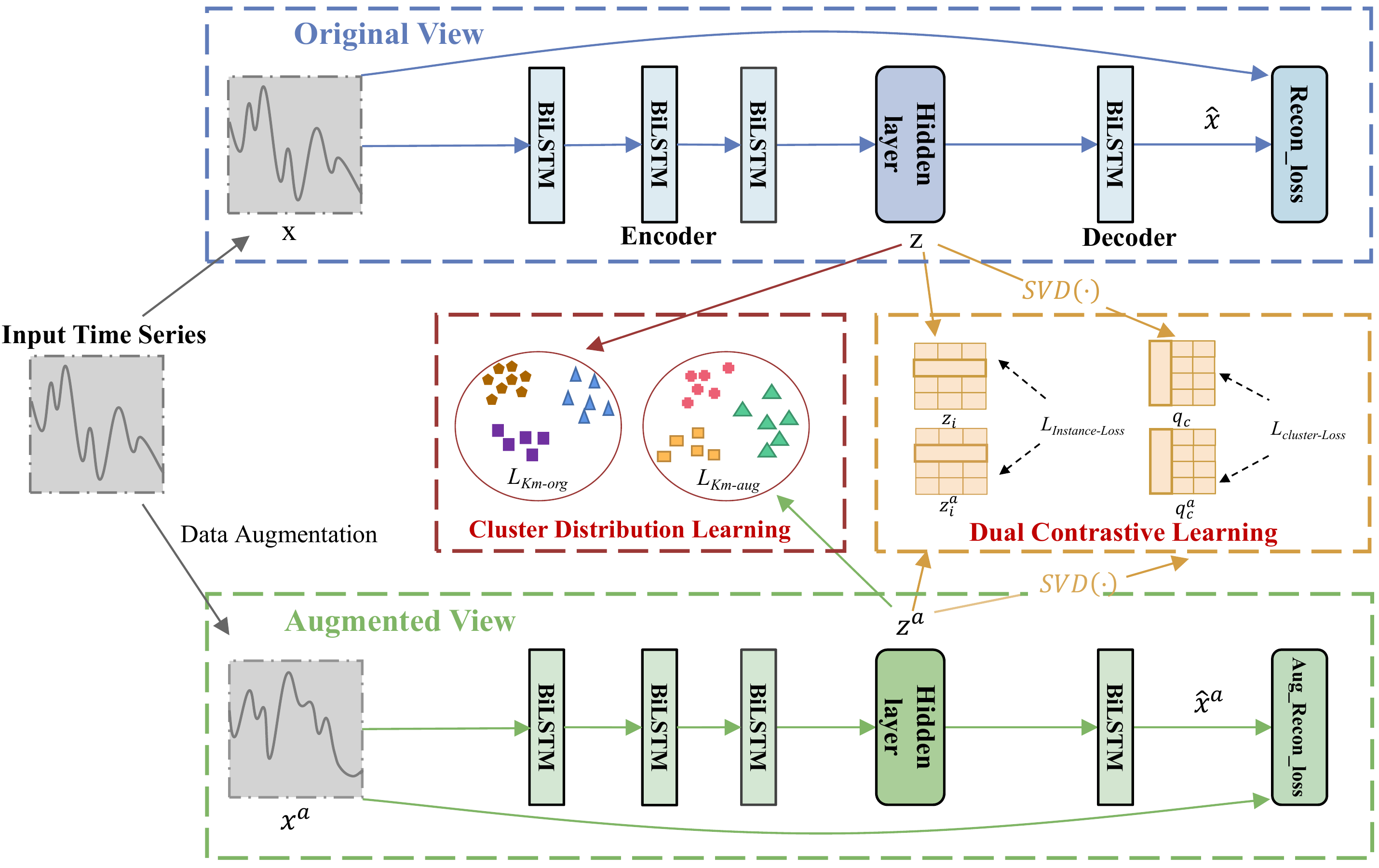}}
 		\caption{The overall architecture of our DTCC framework}
 		\label{fig:framework}
 	\end{center}
 \end{figure*}
 
In light of this, this paper proposes a novel deep clustering approach termed deep temporal contrastive clustering (DTCC), which for the first time, to the best of our knowledge, introduces the contrastive learning into the deep time series clustering research. As illustrated in Fig.~\ref{fig:framework}, the data augmentation (randomly selected from a family of augmentations) is performed on each time series sample, leading to two parallel views, namely, the original view (corresponding to the original samples) and the augmented view (corresponding to the augmented samples). Then, two identical temporal auto-encoders are utilized to learn representations for the two views, respectively. To obtain clustering-friendly representations, the soft $k$-means objective is exploited for the cluster distribution learning. Further, two levels of contrastive learning are enforced to jointly leverage the instance-level and cluster-level contrastiveness. With the temporal instance reconstruction (via auto-encoder), the cluster distribution learning, and the dual contrastive learning seamlessly integrated into a unified objective, the overall network can be effectively trained in a self-supervised manner and the time series clustering result can thereby be obtained. Experimental results on a variety of time series datasets demonstrate the superiority of the proposed DTCC approach.

For clarity, the contributions of this work are summarized as follows:

\begin{itemize}
	\item This paper is the first attempt (to our knowledge) to incorporate the contrastive learning into the deep time series clustering process.
	\item A novel time series clustering approach termed DTCC is proposed, which is capable of  enforcing the temporal instance reconstruction, the cluster distribution learning, and the dual contrastive learning simultaneously.
	\item Extensive experiments are conducted on ten time series datasets, which confirm the superior clustering performance of our DTCC approach.
\end{itemize}
    
The rest of the paper is organized as follows. The related works on time series clustering are reviewed in Section~\ref{sec:related work}. The overall framework of DTCC is described in Section~\ref{sec:framework}. Section~\ref{sec:experiments} reports the experimental results. Finally, Section~\ref{sec:conclusion} concludes this paper.

\section{Related work}
\label{sec:related work}
Time series clustering has been an important research topic in time series analysis. In this section, we first review the traditional time series clustering methods, which can be mainly divided into two categories, namely, the raw data based methods and the feature based methods. Then we proceed to review the  recent advances in the deep time series clustering methods.

The raw data based methods are typically applied to the original time series features, which achieve the clustering result by exploiting some modified distance or similarity function (e.g. via scaling and distortion). Petitjean et al. \cite{petitjean2011global} utilized the dynamic time warping (DTW) metric and performed clustering via the computation of the average of a set of sequences. Yang and Leskovec \cite{yang2011patterns} proposed the $k$-Spectral Centroid($k$-SC) method, which explores the temporal dynamics by using a similarity metric that is invariant to scaling and shifting. Paparrizos and Gravano \cite{paparrizos2015k} proposed a $k$-Shape method for preserving the shapes of time series, which can effectively compute the centroids of time series under the scaling and shift invariances.

The feature based methods attempt to extract feature representations of time series for the clustering purpose, which can reduce the impact of noise or outliers when compared with the raw data based methods.  Guo et al. \cite{guo2008time} converted the raw time series data into feature vectors of lower dimensions by independent component analysis (ICA), and then employed a modified $k$-means clustering algorithm upon the extracted feature vectors. Zakaria et al. \cite{zakaria2012clustering} exploited a concept called shapelets to obtain the local patterns in time series for the clustering task. Yang et al. \cite{yang2011l2} performed time series clustering by taking advantage of pseudo-labels, which adopts the local learning for selecting discriminative features. 

In recent years, the deep learning has shown it advantageous ability in simultaneous representation learning and clustering for the time series data, and several deep time series clustering methods have been devised. For example, Madiraju \cite{madiraju2018deep} proposed the deep temporal clustering (DTC) method, which utilizes an auto-encoder to reduce the temporal dimension and incorporates a temporal clustering layer for learning the clustering assignment. Ma et al. \cite{ma2019learning} utilized a seq2seq model to reconstruct the time series and integrated the $k$-means objectives into the model to guide the representation learning and clustering. Further, they introduced a fake-sample generation strategy, and designed an auxiliary classification task to enhance the discriminative ability of the encoder \cite{ma2019learning}. Anand et al.\cite{anand2021delta} dealt with time series clustering based on feature transformation and feature learning, where each 1-D time-series is transformed into a 2-D image sample and the feature set is obtained by utilizing a pre-trained convolutional neural network on the transformed search space. Ma et al. \cite{ma2020self} developed a time series clustering framework called self-supervised time series clustering network (STCN) by jointly optimizing the feature extraction and clustering. To capture the temporal dynamics and maintain the local structures of time series,  a recurrent neural network (RNN) in the feature extraction module is exploited to perform the one-step time series prediction that acts as the reconstruction of input data. Then, model-based dynamic features are obtained by the output layer of the RNN and further fed into a self-supervised clustering module to achieve the predict labels \cite{ma2020self}.

\section{Proposed Framework}
\label{sec:framework}
In this paper, we present a deep clustering approach termed DTCC for time series data, which incorporates the time series reconstruction (via auto-encoder), the cluster distribution learning, and the dual contrastive learning in a unified framework. In the following, Section~\ref{sec:framework_overview} first describes the overall framework of our DTCC approach. Then the time series reconstruction, the cluster distribution learning, and the dual contrastive learning are depicted in Sections~\ref{sec:reconstruction loss}, \ref{sec:K-means loss}, and \ref{sec:contrastive loss}, respectively. Finally, the overall loss function of DTCC is presented in Section~\ref{sec:overall_loss}.

\subsection{Framework Overview}
\label{sec:framework_overview}

The overall framework of the proposed DTCC approach is illustrated in Fig.~\ref{fig:framework}. In the network architecture of DTCC,  we utilize two identical temporal auto-encoders to reconstruct the input time series data in two different views. the cluster distribution learning module is designed to enforce clustering-friendly representation learning. Further, the dual contrastive learning module is incorporated to learn discriminative representations via simultaneously instance-level and cluster-level contrastive learning.

Specifically, besides the original time series, we construct an augmented view by performing time-series-specific augmentations on the input data. Then, the original sample and the augmented sample are respectively encoded into latent temporal representations via an encoder,  which will be exploited in the following clustering task. The encoder is instantiated by a multi-layer dilated recurrent neural network (RNN). The clustering-friendly representations are obtained by concatenating the last hidden state output of each layer in the multi-layer dilated RNN \cite{Chang2017}. Thereby, two identical encoder-decoder (auto-encoder) networks are trained via  the temporal reconstruction loss, while two soft $k$-means objectives are utilized for the cluster distribution learning.

With the cluster distribution learning module that deploys two $k$-means objectives (on the original view and the augmented view, respective) to guide the latent temporal representation learning, the dual contrastive learning module further jointly enforces the instance-level contrastive learning and the cluster-level contrastive learning. With the simultaneous modeling of the temporal reconstruction, the cluster distribution learning, and the dual contrastive learning, the overall network of DTCC can be trained in a self-supervised manner and thus the clustering result can be learned.

\subsection{Temporal Auto-encoder Reconstruction with Parallel Views}
\label{sec:reconstruction loss}
Given a time series dataset $\mathcal{X}_{data}=\{{x}_{1},\ldots,{x}_{i},\ldots,{x}_{N}\}$ with $N$ samples (or instances), where the $i$-th time series sample ${x}_{i} \in {R}^{m}$ is denoted as ${x}_{i}=({x}_{i,1},{x}_{i,2},\ldots,{x}_{i,m})$ and ${x}_{i,t}$ is the corresponding data at the $t$-th time step. Upon the original time series sample ${x}_{i}$, we perform a time series data augmentation $T^{a}$, which is randomly selected from a family of augmentations $\mathcal{T}$, to obtain an augmented sample $x_{i}^{a}=T(x_{i})$. Note that the original sample ${x}_{i}$ and the augmented sample ${x}_{i}^{a}$ are fed to two different views of two identical auto-encoders, respectively. 

In the auto-encoder, we employ a bidirectional multi-layer dilated RNN as the encoder for capturing the dynamics and multi-scale characteristics of time series, and utilize a single-layer RNN as the decoder. Specifically, the encoder is denoted as $G_{e}: {x}_{i} \rightarrow z_{i}$ for the original view, $G_{e}: {x}_{i}^{a} \rightarrow z_{i}^{a}$) for the augmented view,  which non-linearly map the input time series (of both views) to the latent space $\mathcal{Z}$, where ${z}_{i} \in \mathcal{Z}$ and ${z}_{i}^{a} \in \mathcal{Z}$ denote the learned latent representations of the original sample and the augmented sample, respectively. Also, we denote the decoders in the original view and the augmented view as $G_{d}: {z}_{i} \rightarrow \hat{x}_{i}$ and $G_{d}: {z}_{i}^{a} \rightarrow \hat{x}_{i}^{a}$, respectively. Here, $z_{i} \in R^{d}$ (or $z_{i}^{a} \in R^{d}$) is the $d$-dimensional latent representation of an input time series $x_{i}$ (or ${x}_{i}^{a}$). Formally, the encoding process can be denoted as
\begin{align}
z_{i}=G_{e}\left({x}_{i}\right),
\end{align}
\begin{align}
z_{i}^{a}=G_{e}\left({x}_{i}^{a}\right).
\end{align}
Specifically, $z_{i}$(or $z_{i}^{a}$) is the concatenation of the last hidden state of each layer of the multi-layer Dilated RNN. Thereafter,  $z_{i}$ and $z_{i}^{a}$ are fed to two decoder networks for reconstructing the input data, respectively.

For the decoding task, we obtain the output of the two decoders as $\hat{{x}}_{{i}}$ and $\hat{{x}}_{{i}}^{{a}}$, where $\hat{{x}_{{i}}} \in R^{m}$ and $\hat{{x}}_{{i}}^{{a}} \in R^{m}$ are respectively given by

\begin{align}
\hat{{x}}_{{i}}=G_{d}\left({z}_{{i}}\right),
\end{align}
\begin{align}
\hat{{x}}_{{i}}^{{a}}=G_{d}\left({z}_{{i}}^{{a}}\right).
\end{align}

Let $n$ denote the number of samples in a mini-batch. Then the mean square error (MSE) is used to compute the reconstruction losses of the two parallel views (i.e., the original view and the augmented view), that is
\begin{align}
\mathcal{L}_{{recon-org}}=\frac{1}{n} \sum_{i=1}^{n}\left\|{x}_{{i}}-\hat{{x}}_{{i}}\right\|_{2}^{2},
\end{align}
\begin{align}
\mathcal{L}_{{recon-aug}}=\frac{1}{n} \sum_{i=1}^{n}\left\|{x}_{{i}}^{{a}}-\hat{{x}}_{{i}}^{{a}}\right\|_{2}^{2}.
\end{align}

Thereby, the joint reconstruction loss of both views can be represented as
\begin{align}
\mathcal{L}_{{reconsruction}}=\mathcal{L}_{{recon-org}}+\mathcal{L}_{{recon-aug}}
\end{align}

\subsection{Cluster Distribution Learning}
\label{sec:K-means loss}

Though the auto-encoder with the reconstruction losses can learn some informative features from the original time series and its augmented time series, yet it only considers the information reconstruction of each sample, which cannot reflect capture the higher-level (such as the cluster-level) information for tackling the clustering task. In light of this, in this section, we proceed to incorporate the cluster distribution learning in the network architecture.

With the feature representation matrix $Z \in R^{d \times n}$, we seek to minimize a soft $k$-means objective, which can be rewritten as a trace maximization problem associated the Gram matrices ${Z}^{T}{Z}$, and has an optimal local-minima-free global solution \cite{zha2001spectral}. Formally, the $k$-means objective can be converted (via spectral relaxation) into the following form, that is
\begin{align}
\mathcal{L}_{km-{org}}=\operatorname{Tr}\left({Z}^{{T}} {Z}\right)-\operatorname{Tr}\left({Q}^{{T}} {Z}^{{T}} {Z} {Q}\right)
\label{eq8}
\end{align}
where $\operatorname{Tr}(\cdot)$ is the matrix trace, and ${Q} \in R^{n \times k}$ is the cluster indicator matrix. With $Z$ fixed, the $k$-means based clustering distribution loss (for the original view) can be transformed into a trace maximization problem, that is
\begin{align}
\label{eq:max}
\max _{{Q}} \operatorname{Tr}\left({Q}^{{T}} {Z}^{{T}} {Z} {Q}\right), \text { s.t. } {Q}^{{T}} {Q}={I}
\end{align}
For the augmented view, the clustering distribution loss can be calculated as follows:
\begin{align}
\mathcal{L}_{km-{aug}}=\operatorname{Tr}\left({(Z^a)}^{{T}}Z^a\right)-\operatorname{Tr}\left({(Q^a)}^{{T}} {(Z^a)}^{{T}} {Z^a} {Q^a}\right)
\label{eq10}
\end{align}
Similarly, it can be transformed into a trace maximization problem, that is
\begin{align}
\label{eq:max_aug}
\max _{{Q^a}} \operatorname{Tr}\left({(Q^a)}^{{T}} {(Z^a)}^{{T}} {Z^a} {Q^a}\right), \text { s.t. } {(Q^a)}^{{T}} {Q^a}={I}
\end{align}
Thereby, the cluster distribution loss for both views can be written as
\begin{align}
	\mathcal{L}_{cd} = \frac{1}{2}(\mathcal{L}_{km-{org}}+\mathcal{L}_{km-{aug}}).
\end{align}
In this paper, the cluster distribution learning is coupled with instance reconstruction and contrastive learning. A regularization term $\lambda$ is introduced for learning feature representations with tight clustering structures. Thereby, the objective function can be represented as
\begin{equation}
\begin{aligned}
&\min _{{Z}, {Q},{Z^a}, {Q^a}} J({Z})+\lambda \mathcal{L}_{cd},\\
&\text { s.t. } {Q}^{{T}} {Q}={I},{(Q^a)}^{{T}} {Q^a}={I}
\end{aligned}
\end{equation}
where $J({Z})$ is the sum of the two reconstruction losses and the two contrastive losses. In the following, the contrastive losses (for contrastive learning) will be described in Section~\ref{sec:contrastive loss}. 

For the original view, the representation ${Z}$ is dynamically learned by the auto-encoder as well as the training process of the network in DTCC, where ${Z}$ and ${Q}$ are jointly optimized. The same network update as above is performed for the augmented version. Thus, we focus on the network updating of the original view, and divide it into the following two steps.

First, with ${Q}$ fixed, the standard stochastic gradient descent (SGD) is adopted to update ${Z}$, with the gradient given as $\nabla J({Z})+\lambda {Z}\left({I}-{Q} {Q}^{{T}}\right)$.

Second, with ${Z}$ fixed, according to the Ky Fan theorem, we update ${Q}$ by calculating the $k$-truncated singular value decomposition (SVD) of $Z$. In this paper, it is updated every five epochs so as to avoid potential instability. Thereby, the cluster distribution learning via the $k$-means objective can guide the temporal representation learning with better cluster structures.

\subsection{Dual Contrastive Learning}
\label{sec:contrastive loss}

In the previous section, the cluster distribution learning is performed on the original view and the augmented view separately. In this section, we proceed to bridge the original view and the augmented view via two levels of contrastive learning. In the following, the instance-level contrastive learning (via the instance contrastive loss) and the cluster-level contrastive learning (via the cluster contrastive loss) will be described in Sections~\ref{sec:instance_contrastive} and \ref{sec:cluster_contrastive}, respectively.

\subsubsection{Instance Contrastive Loss}
\label{sec:instance_contrastive}
The instance-level contrastive learning aims to maximize the similarity between the learned representations of an original sample and its augmented sample, while minimizing the similarity between the representations of different samples. Specifically, DTCC takes the representations of the same input time series sample from the two views as the positive sample pairs, and the representations of different input time series samples as the negative sample pairs. 

Let $z_{i}$ and $z_{i}^a$ denote the representations for the $i$-th time series sample from the original view and the augmented view, respectively. The instance contrastive loss for $z_{i}$ can be written as
\begin{align}
{\ell}_{z_{i}}=-\log \frac{\exp \left(sim\left(z_{i}, z_{i}^{a}\right) / \tau_{I}\right)}{\sum_{j=1}^{n}\left[\exp \left(sim\left(z_{i}, z_{j}\right) / \tau_{I}\right)+\mathbbm{1}_{[i \neq j]}\exp \left(sim\left(z_{i}, z_{j}^{a}\right) / \tau_{I}\right)\right]}
\end{align}
where $\tau_{I}$ is the instance-wise temperature parameter, $n$ is the batch size, $\mathbbm{1}$ is the indicator function, and $sim(u,v)={u}^{T}{v} /\|{u}\|\|{v}\|$ is the dot product between $\ell_{2}$-normalized $u$ and $v$. The instance contrastive loss for ${z_{i}^{a}}$ can be similarly defined. Therefore, the instance-wise contrastive loss (for the $n$ samples in a mini-batch) with both view considered can be formulated as
\begin{align}
\mathcal{L}_{{instance }}=\frac{1}{2 n} \sum_{i=1}^{n}({\ell}_{z_{i}}+{\ell}_{z_{i}^{a}})
\label{eq:instance-loss}
\end{align}

\subsubsection{Cluster Contrastive Loss}
\label{sec:cluster_contrastive}

Besides the instance-level contrastive learning, we further enforce the cluster-level contrastive learning, which constrains the \emph{consistency} between the cluster distributions of the two views. 
According to Eq.(\ref{eq8}) and Eq.(\ref{eq10}), let $Q \in R^{n \times k}$ and $Q^{a} \in R^{n \times k}$ denote the cluster indicator matrices for a mini-batch from the two views, where $k$ is the number of clusters. Note that $Q$ and $Q^{a}$ are obtained by computing the $k$-truncated SVD of $Z$ and $Z^{a}$, respectively. 

Let $q_{i}$ be the $i$-th column of $Q$, which can be seen as a representation of the $i$-th cluster (with its $j$-th entry indicating the probability that the $j$-th sample belong to this cluster). To enable the cluster-level contrastive learning, the two clusters $q_{i}$ with $q_i^{a}$ from two different views are regarded as a positive cluster pair $\{q_i,q_i^{a}\}$, where $q_i^{a}$ denotes the $i$-th column of $Q^{a}$. Meanwhile, the remaining $(2k-2)$ pairs within the same mini-batch are regarded as negative cluster pairs. Thereby, the cluster contrastive loss for cluster $q_{i}$ can be formulated as
\begin{align}
\ell_{q_{i}}=-\log \frac{\exp(sim(q_{i}, q_{i}^{a}) / \tau_{C})}{\sum_{j=1}^{k}[\exp(sim(q_{i},q_{j}) / \tau_{C})+\mathbbm{1}_{[i \neq j]}\exp (sim(q_{i}, q_{j}^{a}) / \tau_{C})]}
\end{align}
where $\tau_{C}$ is the cluster-wise temperature parameter. Here, we utilize the cosine similarity to measure the similarity between cluster pairs, that is
\begin{align}
sim\left(q_{i}, q_{i}^{a}\right)=\frac{\left(q_{i}\right)^{\top}\left(q_{i}^{a}\right)}{\left\|q_{i}\right\|\left\|q_{i}^{a}\right\|}
\end{align}
Then,  with $k$ clusters on each view, the cluster-wise contrastive loss for the totally $2k$ clusters on the two views can be defined as
\begin{align}
\begin{split}
\mathcal{L}_{cluster}=\frac{1}{2 k} \sum_{i=1}^{k}({\ell}_{q_{i}}+\ell_{q_{i}^{a}})-{{H}(q)}
\label{eq:cluster-loss}
\end{split}
\end{align}
where $H(q)=-\sum_{i=1}^{k}\left[P\left(q_{i}\right) \log P\left(q_{i}\right)+P\left(q_{i}^{a}\right) \log P\left(q_{i}^{a}\right)\right]$ is the entropy of the cluster-assignment probability, $P\left(q_{i}\right)=\frac{1}{n}\sum_{j=1}^n q_{ji}$, $P\left(q_{i}^{a}\right)=\frac{1}{n}\sum_{j=1}^n q_{ji}^a$, and $q_{ji}$ and $q_{ji}^a$ are respectively the $(j,i)$-th entry of $Q$ and $Q^{a}$. 
The entropy term is incorporated to avoid the degenerate solution that most instances are assigned to the same cluster.

\subsection{Overall Loss Function}
\label{sec:overall_loss}
Finally, with the temporal reconstruction via auto-encoders, the cluster distribution learning, and the the dual contrastive learning, the overall loss function of our DTCC framework is formulated as
\begin{align}
\begin{split}
\label{eq:total-loss}
\mathcal{L}_{D T C C}=\mathcal{L}_{{reconstruction}}+
\mathcal{L}_{{instance }}+\mathcal{L}_{{cluster }}+\lambda\mathcal{L}_{cd}.
\end{split}
\end{align}
Thereby, the network architecture of DTCC can be trained in a self-supervised manner. For clarity, the overall process of the proposed DTCC method is summarized in Algorithm~\ref{alg:alg}.

\begin{algorithm}[h]
 \label{alg:alg}
 \caption{Deep Temporal Contrastive Clustering (DTCC)}
 \LinesNumbered
 \KwIn{Data set: $\mathcal{X}_{data}$; Number of clusters: $k$; Batch size: $n$; Alternate: $T$; Training epochs: $E$.}
 \KwOut{The clustering result $Q$}
 \For{epoch=$1$ to $E$}{
 Sample a mini-batch of time series $\{x_i\}_{i=1}^{n}$ from $\mathcal{X}_{data}$.\\

  For a mini-batch of time series $\{x_i\}_{i=1}^{n}$, generate the augmented samples.\\

  Update the latent representations $\left\{{z}_{i}=G_{e}\left(x_{i}\right)\right\}_{i=1}^{n}$  and $\left\{{z}_{i}^{a}=G_{e}\left({x}_{i}^a\right)\right\}_{i=1}^{n}$ by minizing the overall loss $\mathcal{L}_{DTCC}$ with SGD.\\
  \If {$epoch \% T=0$}{
  Update $\{q_i\}_{i=1}^{n}$ using the closed-form solution of Eq.(\ref{eq:max})\\
  Update$\{q_i^a\}_{i=1}^{n}$ using the closed-form solution of Eq.(\ref{eq:max_aug})\\
  }
 }
Obtain the clustering result Q.
\end{algorithm}

\section{Experiments}

\label{sec:experiments}
\subsection{Experimental Settings and Datasets}
In this paper, two identical encoders with bidirectional multi-layer Dilated RNN are utilized to capture the multi-scale features of the time series data, and a single-layer RNN is utilized as  the decoder. The number of units in each layer is set to $\left[n_{1}, n_{2}, n_{3}\right] \subset\{[100,50,50],[50,30,30]\}$. The number of hidden dimensions is $\left(n_{1}+n_{2}+n_{3}\right) \times 2$. In our experiments, we set the fixed number of layers in the network and the number of each dilation layer to 3 and 1, and 4 and 16, respectively. The decoder treats the final state of the encoder as the initial state and makes iterative predictions. The batch size is one half of the sample size. Our experiments are conducted on the Tensorflow platform using a GeForce GTX 1030 4G GPU. The Adam optimizer is applied with an initial learning rate of 5e-3. To evaluate the clustering performance, two evaluation metrics are adopted, namely, the normalized mutual information (NMI) \cite{Huang2021} and the Rand index (RI) \cite{rand1971objective}. 
 
In the experiments, ten real-world  time series datasets from the UCR Archive \cite{dau2019ucr} are used, namely, Beef, dist.phal.outl.agegroup, ECG200, ECGFiveDays, Meat, MoteStrain, OSULeaf, Plane, Prox.phal.outl.ageGroup, and Prox.phal.TW. The detailed statistics of the datasets are given in Table~\ref{table:dataset}. 

\begin{table*}[!t] \vskip 0.1 in
\centering
\caption{Statistics of the used time series datasets}\vskip 0.05 in
\label{table:dataset}
\setlength{\tabcolsep}{4.5mm}{
\begin{tabular}{lllll}
\toprule
Dataset                 & Train & Test & Length & classes \\
\midrule
Beef                    & 30    & 30   & 471    & 5       \\
dist.phal.outl.agegroup & 139   & 400  & 81     & 3       \\
ECG200                  & 100   & 100  & 97     & 2       \\
ECGFiveDays             & 23    & 86  1  & 137    & 2       \\
Meat                    & 60    & 60   & 449    & 3       \\
MoteStrain              & 20    & 1252 & 85     & 2       \\
OSULeaf                 & 200   & 242  & 428    & 6       \\
Plane                   & 105   & 105  & 145    & 7       \\
Prox.phal.outl.ageGroup & 400   & 205  & 81     & 3       \\
Prox.phal.TW            & 205   & 400  & 81     & 6\\
\bottomrule
\end{tabular}}\vskip 0.1 in
\end{table*}

\subsection{Compared with State-of-the-Art Methods}
In the experiments, we compare our DTCC method against twelve deep or non-deep clustering methods, which are introduced below.

\begin{enumerate}
	\item \textbf{$k$-means} \cite{hartigan1979algorithm} : Perform the $k$-means clustering on the entire time series.
	\item \textbf{UDFS} \cite{yang2011l2}: The unsupervised discriminative feature selection method exploits the manifold structure, the local discriminative information, and the feature correlations for selecting discriminative features.
	\item \textbf{NDFS} \cite{li2012unsupervised}: The nonnegative discriminant feature selection (NDFS) selects the discriminative features by exploiting the  clustering labels acquired by spectral clustering and the feature selection matrix.
	\item \textbf{RUFS} \cite{qian2013robust}: The robust unsupervised feature selection (RUFS) utilized the local learning regularized robust nonnegative matrix factorization to peform the feature learning.
	\item  \textbf{RSFS} \cite{shi2014robust}: The robust spectral learning for unsupervised feature selection (RSFS) method jointly improves the robustness of graph embedding and sparse spectral regression.
	\item \textbf{KSC}  \cite{yang2011patterns}: The $k$-spectral centroid (KSC) clustering method performs clustering by identifying the cluster centroids with the time series similarity measure.
	\item \textbf{$k$-DBA)} \cite{petitjean2011global} : The $k$-DBA method utilizes the $k$-means and the dynamic time warping distance to acquire the clustering result.
	\item \textbf{u-shapelet} \cite{zakaria2012clustering}: The u-shapelet method performs time series clustering by exploiting some local patterns and ignoring the rest of the data.	
	\item  \textbf{DTC} \cite{madiraju2018deep}: The deep temporal clustering (DTC) method naturally integrates the dimensionality reduction and the temporal clustering into an end-to-end unsupervised learning model.
	\item \textbf{DEC} \cite{Xie2016}: The deep embedding clustering (DEC) method can jointly learn the feature representations and the cluster assignments by optimizing a KL-divergence based clustering objective.
	\item \textbf{DTCR} \cite{ma2019learning}:
	The deep temporal clustering representation (DTCR) method incorporates the auto-encoder reconstruction and a $k$-means objective into the seq2seq model for learning cluster-specific temporal representations.
	\item \textbf{CRLI} \cite{ma2021learning}: The clustering representation learning on incomplete time-series data (CRLI) method simultaneously optimizes the imputation and the clustering process in a unified deep learning framework. 
\end{enumerate}

\begin{table*}[!t]\vskip 0.1 in
	\caption{The NMI scores of different clustering methods on the ten time series datasets} \vskip 0.05 in
	\label{table:NMI}
	\scriptsize
	\renewcommand\arraystretch{1.25} 
	\setlength{\tabcolsep}{1.185mm}{
		\begin{tabular}{lllllllllllllll}
			\toprule
			Dataset                 & K-means           & UDFS            & NDFS            &RUFS   & RSFS            & KSC    & KDBA              \\
			\midrule
			Beef                    & 0.2925            & 0.2718          & 0.3647          & \multicolumn{1}{c}{0.3799} & 0.3597          & \multicolumn{1}{c}{0.3828} & 0.3340            \\
			dist.phal.outl.agegroup & 0.1880            & 0.3262          & 0.1943          & \multicolumn{1}{c}{0.2762} & 0.3548          & \multicolumn{1}{c}{0.3331} & 0.4261            \\
			ECG200                  & 0.1403            & 0.1854          & 0.1403          & \multicolumn{1}{c}{0.2668} & \textbf{0.2918} & \multicolumn{1}{c}{0.1403} & 0.1886            \\
			ECGFiveDays             & 0.0002            & 0.0600          & 0.1296          & \multicolumn{1}{c}{0.0352} & 0.1760          & \multicolumn{1}{c}{0.0682} & \textbf{0.1983}   \\
			Meat                    & 0.2510            & 0.2832          & 0.2416          & \multicolumn{1}{c}{0.1943} & 0.3016          & \multicolumn{1}{c}{0.2846} & 0.3661            \\
			MoteStrain              & 0.0551            & 0.1187          & 0.1919          & \multicolumn{1}{c}{0.1264} & 0.2373          & \multicolumn{1}{c}{0.3002} & 0.0970            \\
			OSULeaf                 & 0.0208            & 0.0200          & 0.0552          & \multicolumn{1}{c}{0.0246} & 0.0463          & \multicolumn{1}{c}{0.0421} & 0.0327            \\
			Plane                   & 0.8598            & 0.8046          & 0.8414          & \multicolumn{1}{c}{0.8675} & 0.8736          & \multicolumn{1}{c}{0.9218} & 0.8784            \\
			Prox.phal.outl.ageGroup & 0.0635            & 0.0182          & 0.0830          & \multicolumn{1}{c}{0.0726} & 0.0938          & \multicolumn{1}{c}{0.0682} & 0.0377            \\
			Prox.phal.TW            & 0.0082            & 0.0308          & 0.2215          & \multicolumn{1}{c}{0.1187} & 0.0809          & \multicolumn{1}{c}{0.1919} & 0.2167            \\
			\midrule
			Avg. NMI                 & 0.1879            & 0.2119          & 0.2464          & \multicolumn{1}{c}{0.2362} & 0.2816          & \multicolumn{1}{c}{0.2732} & 0.2776            \\
			\bottomrule
			\toprule
			Dataset                 & u-shapelet        & DTC             & DEC             & DTCR            & CRLI                       &\multicolumn{2}{c}{DTCC}              \\
			\midrule
			Beef                    & 0.3413            & 0.2751          & 0.2463          & 0.4484 & \textbf{0.4570 }                    & \multicolumn{2}{c}{\textbf{[0.4753]}} \\
			dist.phal.outl.agegroup & 0.2577            & 0.3406          & \textbf{0.4405}                     & 0.3957          & 0.3080                     & \multicolumn{2}{c}{\textbf{[0.4596]}} \\
			ECG200                  & 0.1323            & 0.0918          & 0.1885          & 0.1805          & 0.2064                     & \multicolumn{2}{c}{\textbf{[0.3009]}}   \\
			ECGFiveDays             & 0.1498            & 0.0022          & 0.0178          & 0.0900          & 0.1875                     & \multicolumn{2}{c}{\textbf{[0.3623]}} \\
			Meat                    & 0.2716            & 0.2250          & 0.5176          & 0.3987          & \textbf{[0.6608]}          & \multicolumn{2}{c}{\textbf{0.6099}} \\
			MoteStrain              & 0.0082            & 0.0094          & \textbf{0.3867} & 0.3218          & 0.2748                     & \multicolumn{2}{c}{\textbf{[0.4653]}} \\
			OSULeaf                 & 0.0203            & \textbf{0.2201}          & 0.2141          & 0.1953          & 0.2071                     & \multicolumn{2}{c}{\textbf{[0.2354]}}   \\
			Plane                   & \textbf{[1.0000]} & 0.8678          & 0.8947          & 0.9049          & \textbf{[1.0000]}          & \multicolumn{2}{c}{\textbf{0.9252}}   \\
			Prox.phal.outl.ageGroup & 0.0332            & 0.4153          & 0.2500          & 0.4878          & \textbf{0.5227}                     & \multicolumn{2}{c}{\textbf{[0.5317]}} \\
			Prox.phal.TW            & 0.0107            & \textbf{0.6199} & 0.5864          & 0.5190          & 0.5709                     & \multicolumn{2}{c}{\textbf{[0.6213]}} \\
			\midrule
			Avg. NMI                 & 0.2225            & 0.3067          & 0.3743           &0.3942 & \textbf{0.4395}                    & \multicolumn{2}{c}{\textbf{[0.4987]}}\\
			\bottomrule
	\end{tabular}}
\end{table*}

\begin{figure*}[!t]\vskip 0.1 in
	\centering
	{\subfigure[]{
			\includegraphics[width=0.48\textwidth]{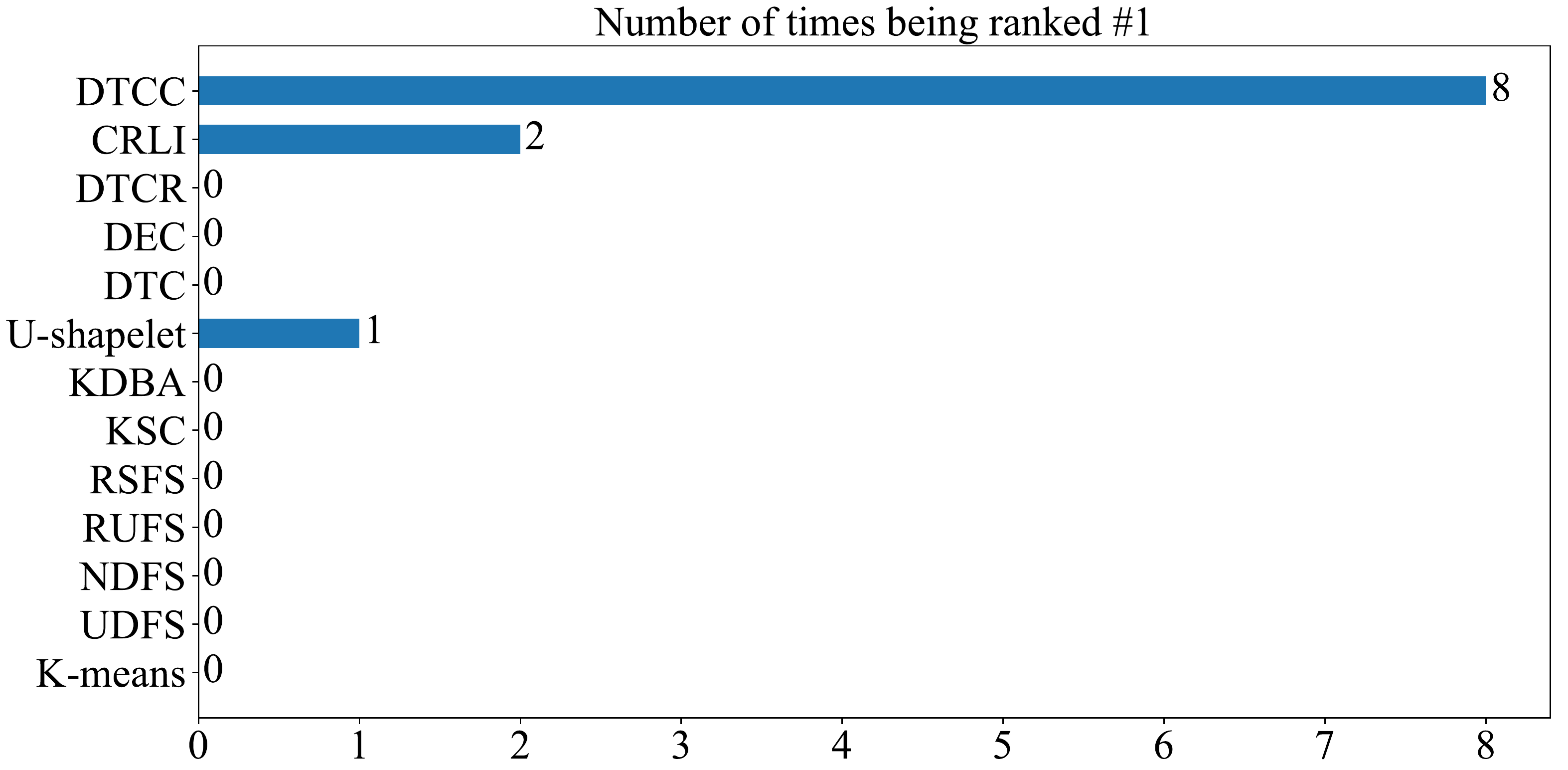}\label{fig:rank1nmi}}}
	{\subfigure[]{
			\includegraphics[width=0.48\textwidth]{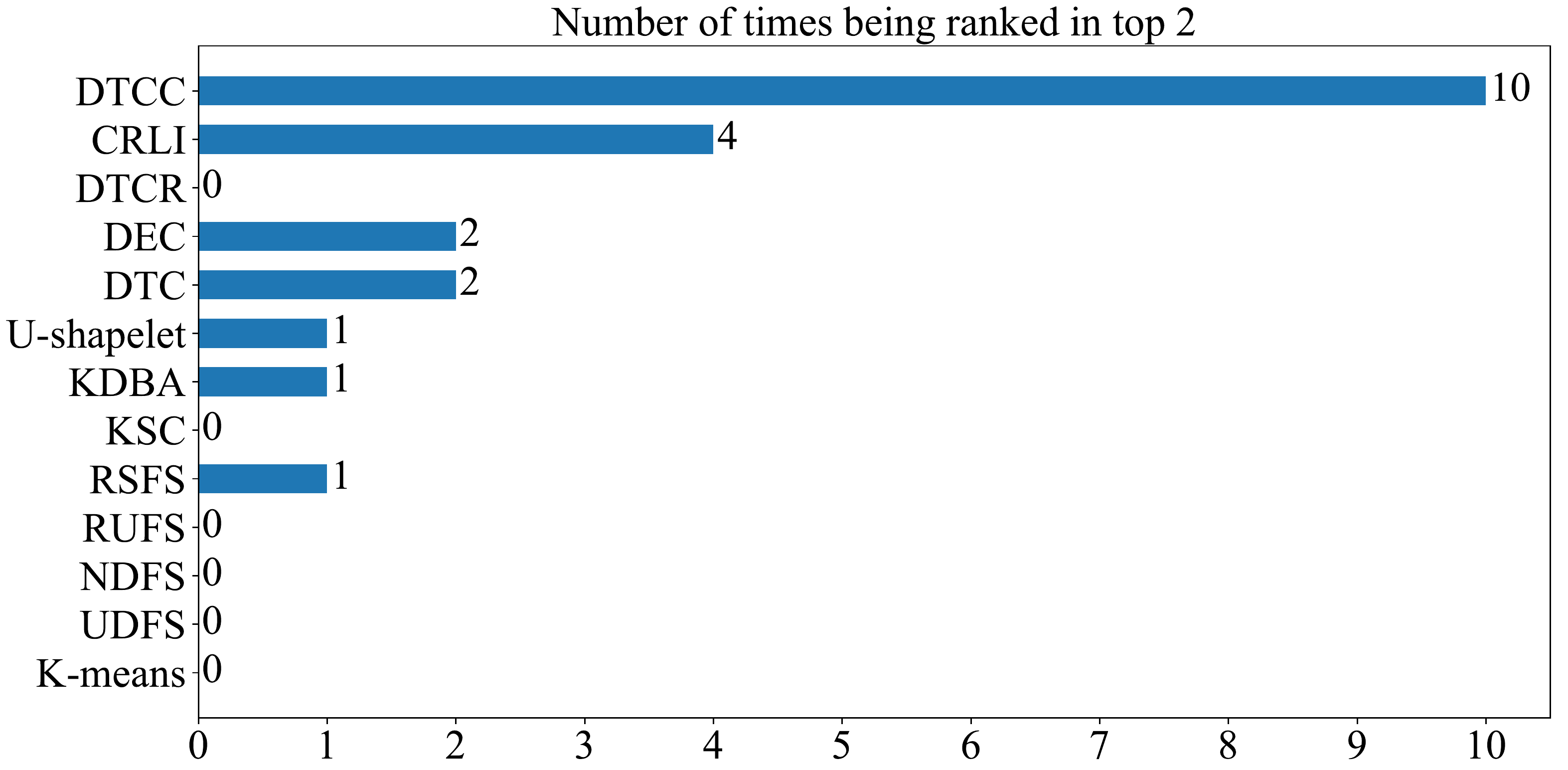}\label{fig:rank2nmi}}}\vskip -0.1 in
	\caption{The number of times being ranked in top 1 or top 2 on the ten datasets (w.r.t. the results in Table~\ref{table:NMI}).}\label{ranknmi}
\end{figure*}

\begin{table*}[!t] \vskip 0.1 in
	\caption{The RI scores of different clustering methods on the ten time series datasets} \vskip 0.05 in
	\label{table:RI}
	\scriptsize
	\renewcommand\arraystretch{1.25} 
	\setlength{\tabcolsep}{1.15mm}{
		\begin{tabular}{lllllllllllllll}
			\toprule
			Dataset                 & K-means           & UDFS              & NDFS            &RUFS                        & RSFS            & KSC              & KDBA              \\
			\midrule
			Beef                    & 0.6713            & 0.6759            & 0.7034          & 0.7149                     & 0.6975          & 0.7057           & 0.6713            \\
			dist.phal.outl.agegroup & 0.6171            & 0.6531            & 0.6239          & 0.6252                     & 0.6539          & 0.6535           & 0.6750            \\
			ECG200                  & 0.6315            & 0.6533            & 0.6315          & \textbf{[0.7018]}          & \textbf{0.6916} & 0.6315           & 0.6018            \\
			ECGFiveDays             & 0.4783            & 0.5020            & 0.5573          & 0.5020                     & 0.5953          & 0.5257           & 0.5573            \\
			Meat                    & 0.6595            & 0.6483            & 0.6635          & 0.6578                     & 0.6657          & 0.6723           & 0.6816            \\
			MoteStrain              & 0.4947            & 0.5579            & 0.6053          & 0.5579                     & 0.6168          & 0.6632           & 0.4789            \\
			OSULeaf                 & 0.5615            & 0.5372            & 0.5622          & 0.5497                     & 0.5665          & 0.5714           & 0.5541            \\
			Plane                   & 0.9081            & 0.8949            & 0.8954          & 0.9220                     & 0.9314          &\textbf{0.9603}   & 0.9225    \\
			Prox.phal.outl.ageGroup & 0.5288            & 0.4997            & 0.5463          & 0.5780                     & 0.5384          & 0.5305           & 0.5192            \\
			Prox.phal.TW            & 0.4789            & 0.4947            & 0.6053          & 0.5579                     & 0.5211          & 0.6053           & 0.5211            \\
			\midrule
			Avg. RI                  & 0.6030            & 0.6117            & 0.6394          & 0.6367                     & 0.6478          & 0.6519           & 0.6183            \\
			\bottomrule
			\toprule
			Dataset                 & u-shapelet        & DTC               & DEC             & DTCR            & CRLI             & \multicolumn{2}{c}{DTCC}              \\
			\midrule
			Beef                    & 0.6966            & 0.6345            & 0.5954          &0.7452           &\textbf{[0.7655]}  &\multicolumn{2}{c}{\textbf{0.7633}} \\
			dist.phal.outl.agegroup & 0.6273            & \textbf{[0.7812]} & \textbf{0.7785}          & 0.7417          & 0.6520 & \multicolumn{2}{c}{0.7516}            \\
			ECG200                  & 0.5758            & 0.6018            & 0.6422          & 0.6085          & 0.6315 & \multicolumn{2}{c}{0.6733}            \\
			ECGFiveDays             & 0.5968            & 0.5016            & 0.5103          & 0.5419          & \textbf{0.6237}   & \multicolumn{2}{c}{\textbf{[0.6581]}} \\
			Meat                    & 0.6742            & 0.3220            & 0.6475          & 0.6934          & \textbf{[0.8305]} & \multicolumn{2}{c}{\textbf{0.7200}} \\
			MoteStrain              & 0.4789            & 0.5062            & \textbf{0.7435} & 0.6956          & 0.6792            & \multicolumn{2}{c}{\textbf{[0.7740]}} \\
			OSULeaf                 & 0.5525            & 0.7329            & 0.7484          & \textbf{0.7497}          & 0.7476            & \multicolumn{2}{c}{\textbf{[0.7546]}}   \\
			Plane                   & \textbf{[1.0000]} & 0.9040            & 0.9447          & 0.9404          & \textbf{[1.0000]} & \multicolumn{2}{c}{0.9373}            \\
			Prox.phal.outl.ageGroup & 0.5206            & 0.7430            & 0.4263          & \textbf{0.8023} & 0.8020            & \multicolumn{2}{c}{\textbf{[0.8091]}} \\
			Prox.phal.TW            & 0.4789            & 0.8380            & 0.8189          & \textbf{[0.8617]} & 0.8439            & \multicolumn{2}{c}{\textbf{0.8606}}            \\
			\midrule
			Avg. RI                  & 0.6202            & 0.6565            & 0.6856          & 0.7380          & \textbf{0.7576}   & \multicolumn{2}{c}{\textbf{[0.7729]}}\\
			\bottomrule
	\end{tabular}}
\end{table*}

\begin{figure*}[!t]\vskip 0.1 in
	\centering
	{\subfigure[]{\includegraphics[width=0.48\textwidth]{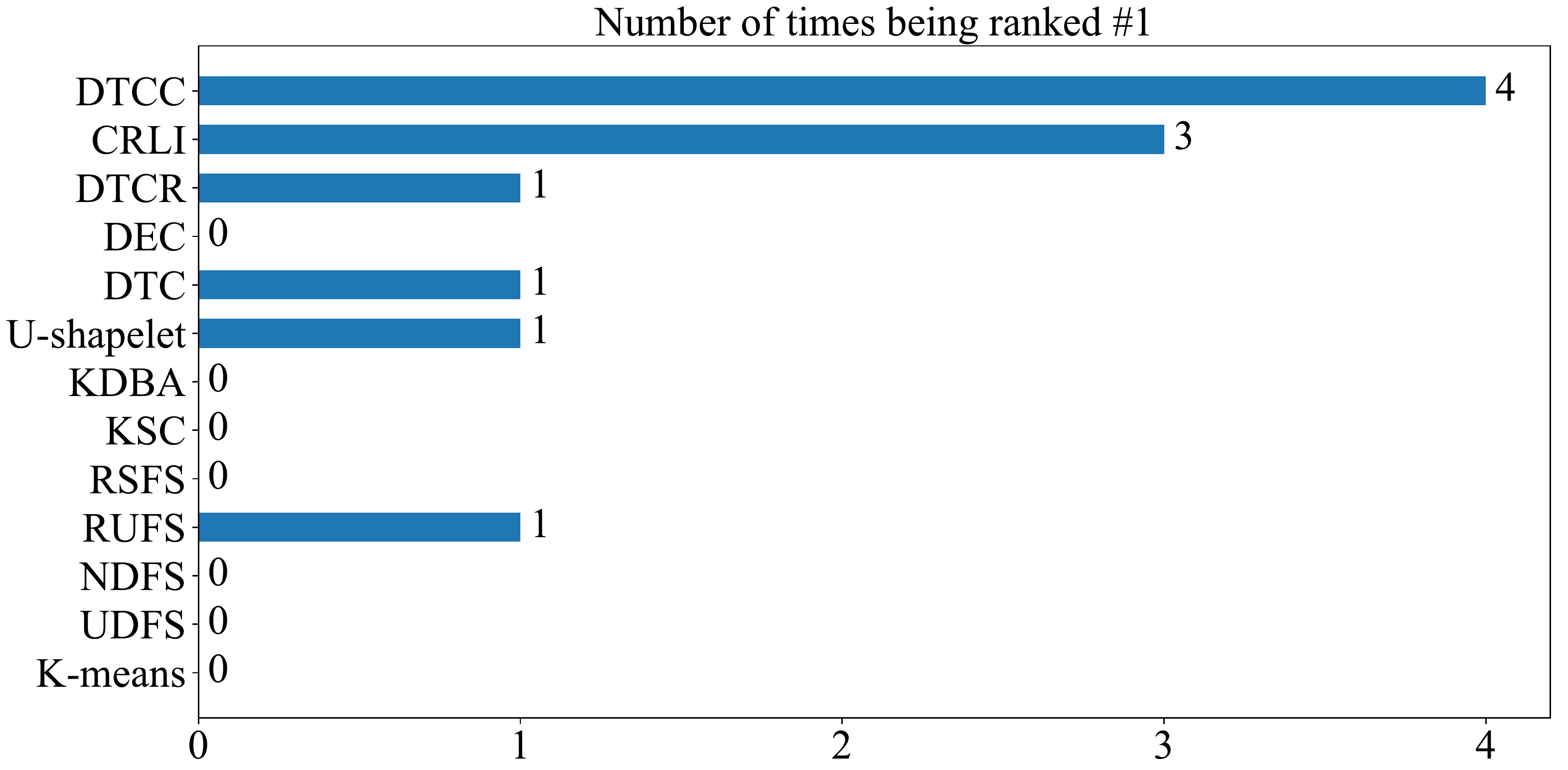}\label{fig:rank1ri}}}
	{\subfigure[]{
			\includegraphics[width=0.48\textwidth]{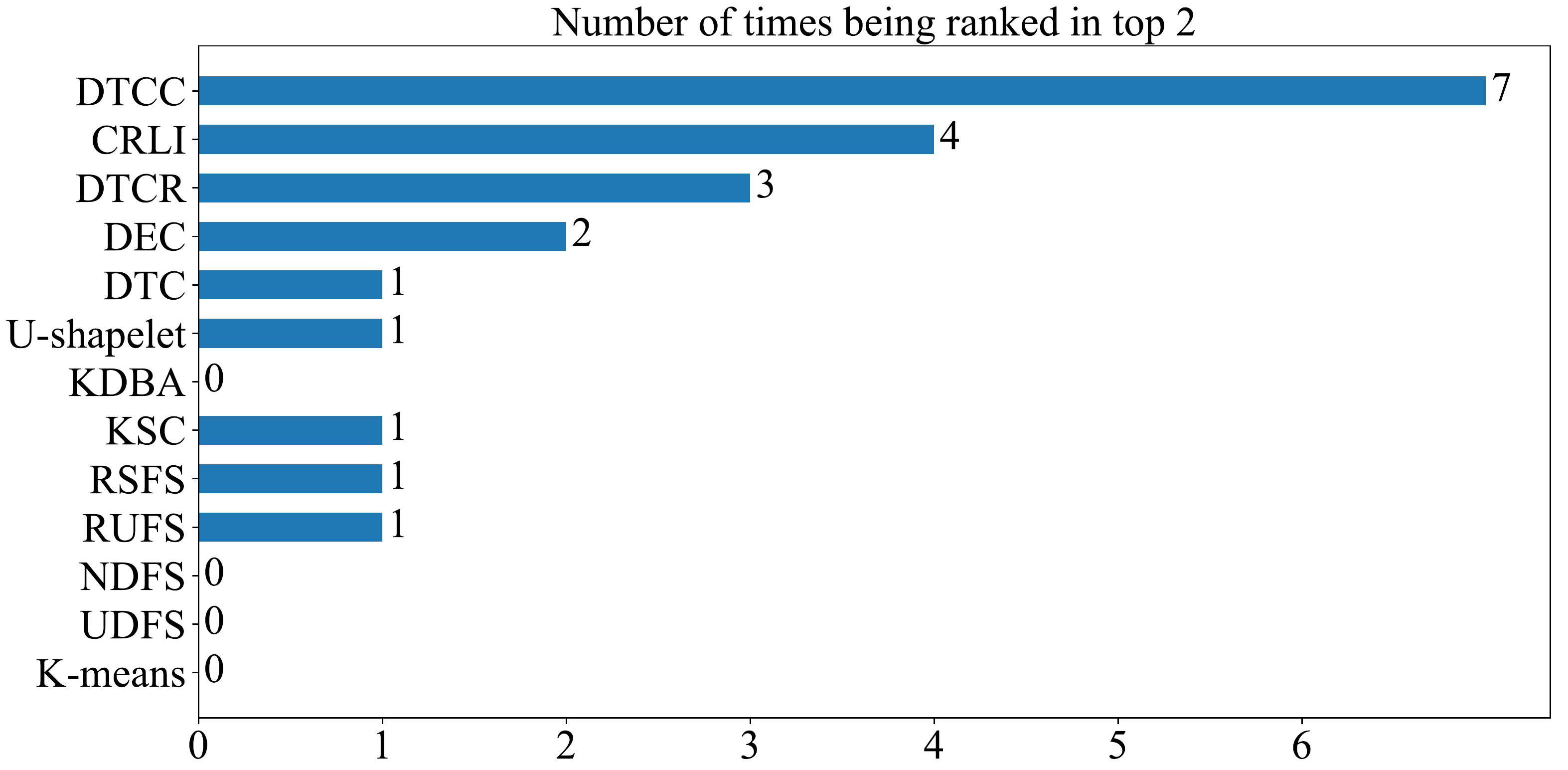}\label{fig:rank2ri}}}
	\vskip -0.1 in
	\caption{The number of times being ranked in top 1 or top 2 on the ten datasets (w.r.t. the results in Table~\ref{table:RI}).}
	\label{rankri}
\end{figure*}

Specifically, the clustering performances (w.r.t. NMI and RI) of our DTCC method and the baseline methods are reported in Tables~\ref{table:NMI} and \ref{table:RI}, respectively. In terms of NMI, as shown in Table~\ref{table:NMI}, the proposed DTCC method achieves the best clustering performance on the Beef, dist.phal.outl.agegroup, ECG200, ECGFiveDays, MoteStrain, OSULeaf, Prox.phal.TW datasets, and Prox.phal.TW datasets. On the other datasets, DTCC is still among the best two clustering methods w.r.t. the NMI scores. The average NMI scores (across the ten datasets) of different methods are also provided in Table~\ref{table:NMI}. Our DTCC obtains the average NMI score of 0.4987, which significantly outperforms the second best method, whose average NMI score is only 0.4395.
In terms of RI, similar advantages of DTCC can also be observed. Our DTCC method obtains the average RI score of 0.7729, which is substantially higher than the average score of 0.7576 obtained by the second best method.

\begin{table*}[!t] \vskip 0.15 in
	\caption{Ablation analysis on the dual contrastive losses in DTCC (w.r.t. NMI). \textbf{ICL}: instance contrastive loss. \textbf{CCL}: cluster contrastive loss.}
	\label{table:ablation study of nmi}
	\renewcommand\arraystretch{1.43} \vspace {2mm}
	\scriptsize
	\setlength{\tabcolsep}{0.3mm}{
		\begin{tabular}{|cc|ccccc|}
			\hline
			\multicolumn{1}{|c|}{\multirow{2}{*}{~~ICL~~}} & \multirow{2}{*}{~~CCL~~} & \multicolumn{5}{c|}{Datasets}                                                                                                                                                               \\ \cline{3-7}
			\multicolumn{1}{|c|}{}                               &                               & \multicolumn{1}{c}{Beef}            & \multicolumn{1}{c}{dist.phal.outl.agegroup} & \multicolumn{1}{c}{ECG200}          & \multicolumn{1}{c}{ECGFiveDays}             & Meat            \\ \hline
			\multicolumn{1}{|c|}{\ding{53}}                              & \checkmark                             & \multicolumn{1}{c}{0.4211}          & \multicolumn{1}{c}{0.4096}                  & \multicolumn{1}{c}{0.1140}          & \multicolumn{1}{c}{0.1291}                  & 0.1683          \\ \cline{1-2}
			\multicolumn{1}{|c|}{\checkmark}                              & \ding{53}                             & \multicolumn{1}{c}{\textbf{0.4753}}          & \multicolumn{1}{c}{0.4229}                  & \multicolumn{1}{c}{0.2436}          & \multicolumn{1}{c}{0.2966}                  & 0.3574          \\ \cline{1-2}
			\multicolumn{1}{|c|}{\ding{53}}                              & \ding{53}                             & \multicolumn{1}{c}{0.4284}          & \multicolumn{1}{c}{0.4170}                  & \multicolumn{1}{c}{0.1126}          & \multicolumn{1}{c}{0.0730}                  & 0.2049          \\ \cline{1-2}
			\multicolumn{1}{|c|}{\checkmark}                              & \checkmark                             & \multicolumn{1}{c}{\textbf{0.4753}} & \multicolumn{1}{c}{\textbf{0.4596}}         & \multicolumn{1}{c}{\textbf{0.3009}} & \multicolumn{1}{c}{\textbf{0.3623}}         & \textbf{0.6099} \\ \hline
			\multicolumn{2}{|c|}{}                                                               & \multicolumn{1}{c}{MoteStrain}      & \multicolumn{1}{c}{OSULeaf}                 & \multicolumn{1}{c}{Plane}           & \multicolumn{1}{c}{Prox.phal.outl.ageGroup} & Prox.phal.TW    \\ \hline
			\multicolumn{1}{|c|}{\ding{53}}                              & \checkmark                             & \multicolumn{1}{c}{0.4040}          & \multicolumn{1}{c}{0.1487}                  & \multicolumn{1}{c}{0.7751}          & \multicolumn{1}{c}{0.5310}                  & 0.5334          \\ \cline{1-2}
			\multicolumn{1}{|c|}{\checkmark}                              & \ding{53}                             & \multicolumn{1}{c}{0.4086}          & \multicolumn{1}{c}{\textbf{0.2435}}         & \multicolumn{1}{c}{0.8570}          & \multicolumn{1}{c}{0.5246}                  & 0.5328          \\ \cline{1-2}
			\multicolumn{1}{|c|}{\ding{53}}                              & \ding{53}                             & \multicolumn{1}{c}{0.3751}          & \multicolumn{1}{c}{0.1794}                  & \multicolumn{1}{c}{0.8342}          & \multicolumn{1}{c}{0.5271}                  & 0.5365          \\ \cline{1-2}
			\multicolumn{1}{|c|}{\checkmark}                              & \checkmark                             & \multicolumn{1}{c}{\textbf{0.4653}} & \multicolumn{1}{c}{0.2354}                  & \multicolumn{1}{c}{\textbf{0.9252}} & \multicolumn{1}{c}{\textbf{0.5317}}         & \textbf{0.6213} \\ \hline
	\end{tabular}}
\end{table*}

\begin{table*}[]
	\caption{Ablation analysis on the dual contrastive losses in DTCC (w.r.t.RI). \textbf{ICL}: instance contrastive loss. \textbf{CCL}: cluster contrastive loss.}
	\label{table:ablation study of ri}
	\renewcommand\arraystretch{1.43} \vspace {2mm}
	\scriptsize
	\setlength{\tabcolsep}{0.3mm}{
		\begin{tabular}{|cc|ccccc|}
			\hline
			\multicolumn{1}{|c|}{\multirow{2}{*}{~~ICL~~}} & \multirow{2}{*}{~~CCL~~} & \multicolumn{5}{c|}{Datasets}                                                                                                                                                               \\ \cline{3-7}
			\multicolumn{1}{|c|}{}                               &                               & \multicolumn{1}{c}{Beef}            & \multicolumn{1}{c}{dist.phal.outl.agegroup} & \multicolumn{1}{c}{ECG200}          & \multicolumn{1}{c}{ECGFiveDays}             & Meat            \\ \hline
			\multicolumn{1}{|c|}{\ding{53}}                              & \checkmark                             & \multicolumn{1}{c}{0.7367}          & \multicolumn{1}{c}{0.7505}                  & \multicolumn{1}{c}{0.6085}          & \multicolumn{1}{c}{0.5777}                  & 0.6332          \\ \cline{1-2}
			\multicolumn{1}{|c|}{\checkmark}                              & \ding{53}                             & \multicolumn{1}{c}{0.7593}          & \multicolumn{1}{c}{0.7445}                  & \multicolumn{1}{c}{0.6724}          & \multicolumn{1}{c}{\textbf{0.6903}}         & 0.6954          \\ \cline{1-2}
			\multicolumn{1}{|c|}{\ding{53}}                              & \ding{53}                             & \multicolumn{1}{c}{0.7423}          & \multicolumn{1}{c}{0.7419}                  & \multicolumn{1}{c}{0.5819}          & \multicolumn{1}{c}{0.5127}                  & 0.6376          \\ \cline{1-2}
			\multicolumn{1}{|c|}{\checkmark}                              & \checkmark                             & \multicolumn{1}{c}{\textbf{0.7632}} & \multicolumn{1}{c}{\textbf{0.7516}}         & \multicolumn{1}{c}{\textbf{0.6733}} & \multicolumn{1}{c}{0.6851}                  & \textbf{0.7200} \\ \hline
			\multicolumn{2}{|c|}{}                                                               & \multicolumn{1}{c}{MoteStrain}      & \multicolumn{1}{c}{OSULeaf}                 & \multicolumn{1}{c}{Plane}           & \multicolumn{1}{c}{Prox.phal.outl.ageGroup} & Prox.phal.TW    \\ \hline
			\multicolumn{1}{|c|}{\ding{53}}                              & \checkmark                             & \multicolumn{1}{c}{0.7426}          & \multicolumn{1}{c}{0.7375}                  & \multicolumn{1}{c}{0.9155}          & \multicolumn{1}{c}{0.8042}                  & 0.7927          \\ \cline{1-2}
			\multicolumn{1}{|c|}{\checkmark}                              & \ding{53}                             & \multicolumn{1}{c}{0.7457}          & \multicolumn{1}{c}{\textbf{0.7575}}         & \multicolumn{1}{c}{0.9396}          & \multicolumn{1}{c}{0.8059}                  & \textbf{0.8612} \\ \cline{1-2}
			\multicolumn{1}{|c|}{\ding{53}}                              & \ding{53}                             & \multicolumn{1}{c}{0.7213}          & \multicolumn{1}{c}{0.7417}                  & \multicolumn{1}{c}{0.9283}          & \multicolumn{1}{c}{0.8042}                  & 0.7969          \\ \cline{1-2}
			\multicolumn{1}{|c|}{\checkmark}                              & \checkmark                             & \multicolumn{1}{c}{\textbf{0.7740}} & \multicolumn{1}{c}{0.7546}                  & \multicolumn{1}{c}{\textbf{0.9373}} & \multicolumn{1}{c}{\textbf{0.8091}}         & 0.8606          \\ \hline
	\end{tabular}}\vskip 0.1 in
\end{table*}

Furthermore, Figs.~\ref{ranknmi} and \ref{rankri} illustrate the number of times of different clustering methods to be ranked in top 1 or top 2 in terms of NMI and RI, respectively. As shown in Figs.~\ref{ranknmi} and \ref{rankri}, our DTCC method is ranked in the top 2 on all of the ten datasets in terms of NMI, and on seven out of the ten datasets in terms of RI, which confirm the highly competitive clustering performance of the proposed DTCC method in comparison with the state-of-the-art methods.

\subsection{Ablation Study}
In this section, the ablation study is conducted. Specifically, the influence of the two contrastive losses are evaluated in Section~\ref{sec:test_contrastive}, and the influence of the cluster distribution loss is tested in Section~\ref{sec:test_distribution_loss}.

\subsubsection{Influence of the Two Contrastive Losses}
\label{sec:test_contrastive}
In DTCC, two levels of constrastive learning are enforced, corresponding to two types of contrastive losses, i.e., the instance contrastive loss and the cluster contrastive loss. In this section, we evaluate the influence of these two contrastive losses in DTCC.

As shown in Table~\ref{table:ablation study of nmi}, our DTCC method with the two types of contrastive losses outperforms or significantly outperforms the variant with no contrastive loss (in terms of NMI). When comparing DTCC with both contrastive losses against the variant with only one of them, the joint use of both contrastive losses exhibits overall better NMI score than using only one level of contrastiveness. Similarity results can also be seen in Table~\ref{table:ablation study of ri}, where the advantage of using the dual contrastive losses has also been verified on most of the benchmark datasets.

\begin{table*}[!t] \vskip 0.15 in
	\caption{Ablation analysis on the cluster distribution losses in DTCC (w.r.t. NMI).}
	\label{table:cluster distribution nmi}
	\renewcommand\arraystretch{1.43} \vspace {2mm}
	\scriptsize
	\setlength{\tabcolsep}{0.3mm}{
		\begin{tabular}{|cc|ccccc|}
			\hline
			\multicolumn{1}{|c|}{\multirow{2}{*}{$\mathcal{L}_{km-{org}}$}} & \multirow{2}{*}{$\mathcal{L}_{km-{aug}}$} & \multicolumn{5}{c|}{Datasets}                                                                                                                                                               \\ \cline{3-7}
			\multicolumn{1}{|c|}{}                         &                              & \multicolumn{1}{c}{Beef}            & \multicolumn{1}{c}{dist.phal.outl.agegroup} & \multicolumn{1}{c}{ECG200}          & \multicolumn{1}{c}{ECGFiveDays}             & Meat            \\ \hline
			\multicolumn{1}{|c|}{\ding{53}}                        & \checkmark                            & \multicolumn{1}{c}{0.4507}          & \multicolumn{1}{c}{0.4163}                  & \multicolumn{1}{c}{0.2187}          & \multicolumn{1}{c}{0.1510}                  & 0.3921          \\ \cline{1-2}
			\multicolumn{1}{|c|}{\checkmark}                        & \ding{53}                            & \multicolumn{1}{c}{0.4710}          & \multicolumn{1}{c}{0.4370}                  & \multicolumn{1}{c}{0.2905}          & \multicolumn{1}{c}{0.1482}                  & 0.4424          \\ \cline{1-2}
			\multicolumn{1}{|c|}{\ding{53}}                        & \ding{53}                            & \multicolumn{1}{c}{0.4606}          & \multicolumn{1}{c}{0.4555}                  & \multicolumn{1}{c}{0.2042}          & \multicolumn{1}{c}{0.2183}                  & 0.4224          \\ \cline{1-2}
			\multicolumn{1}{|c|}{\checkmark}                        & \checkmark                            & \multicolumn{1}{c}{\textbf{0.4753}} & \multicolumn{1}{c}{\textbf{0.4596}}         & \multicolumn{1}{c}{\textbf{0.3009}} & \multicolumn{1}{c}{\textbf{0.3623}}         & \textbf{0.6099} \\ \hline
			\multicolumn{2}{|c|}{}                                                        & \multicolumn{1}{c}{MoteStrain}      & \multicolumn{1}{c}{OSULeaf}                 & \multicolumn{1}{c}{Plane}           & \multicolumn{1}{c}{Prox.phal.outl.ageGroup} & Prox.phal.TW    \\ \hline
			\multicolumn{1}{|c|}{\ding{53}}                        & \checkmark                            & \multicolumn{1}{c}{0.3669}          & \multicolumn{1}{c}{0.1940}                  & \multicolumn{1}{c}{0.8200}          & \multicolumn{1}{c}{0.5093}                  & 0.5459          \\ \cline{1-2}
			\multicolumn{1}{|c|}{\checkmark}                        & \ding{53}                            & \multicolumn{1}{c}{0.3864}          & \multicolumn{1}{c}{0.2308}                  & \multicolumn{1}{c}{0.8404}          & \multicolumn{1}{c}{0.5225}                  & 0.5594          \\ \cline{1-2}
			\multicolumn{1}{|c|}{\ding{53}}                        & \ding{53}                            & \multicolumn{1}{c}{0.3826}          & \multicolumn{1}{c}{0.2072}                  & \multicolumn{1}{c}{0.8485}          & \multicolumn{1}{c}{0.5127}                  & 0.5331          \\ \cline{1-2}
			\multicolumn{1}{|c|}{\checkmark}                        & \checkmark                            & \multicolumn{1}{c}{\textbf{0.4653}} & \multicolumn{1}{c}{\textbf{0.2354}}         & \multicolumn{1}{c}{\textbf{0.9252}} & \multicolumn{1}{c}{\textbf{0.5317}}         & \textbf{0.6213} \\ \hline
	\end{tabular}}
\end{table*}

\begin{table*}[]
	\caption{Ablation analysis on the cluster distribution losses in DTCC (w.r.t. RI).}
	\label{table:cluster distribution ri}
	\renewcommand\arraystretch{1.43} \vspace {0.3mm}
	\scriptsize
	\setlength{\tabcolsep}{0.3mm}{
		\begin{tabular}{|cc|ccccc|}
			\hline
			\multicolumn{1}{|c|}{\multirow{2}{*}{$\mathcal{L}_{km-{org}}$}} & \multirow{2}{*}{$\mathcal{L}_{km-{aug}}$} & \multicolumn{5}{c|}{Datasets}                                                                                                                                                               \\ \cline{3-7}
			\multicolumn{1}{|c|}{}                         &                              & \multicolumn{1}{c}{Beef}            & \multicolumn{1}{c}{dist.phal.outl.agegroup} & \multicolumn{1}{c}{ECG200}          & \multicolumn{1}{c}{ECGFiveDays}             & Meat            \\ \hline
			\multicolumn{1}{|c|}{\ding{53}}                        & \checkmark                            & \multicolumn{1}{c}{0.7610}          & \multicolumn{1}{c}{0.7382}                  & \multicolumn{1}{c}{0.6687}          & \multicolumn{1}{c}{0.6003}                  & 0.7102          \\ \cline{1-2}
			\multicolumn{1}{|c|}{\checkmark}                        & \ding{53}                            & \multicolumn{1}{c}{0.7485}          & \multicolumn{1}{c}{0.7513}                  & \multicolumn{1}{c}{\textbf{0.7040}} & \multicolumn{1}{c}{0.5982}                  & 0.7127          \\ \cline{1-2}
			\multicolumn{1}{|c|}{\ding{53}}                        & \ding{53}                            & \multicolumn{1}{c}{0.7553}          & \multicolumn{1}{c}{\textbf{0.7545}}         & \multicolumn{1}{c}{0.6568}          & \multicolumn{1}{c}{0.6430}                  & \textbf{0.7242} \\ \cline{1-2}
			\multicolumn{1}{|c|}{\checkmark}                        & \checkmark                            & \multicolumn{1}{c}{\textbf{0.7632}} & \multicolumn{1}{c}{0.7516}                  & \multicolumn{1}{c}{0.6733}          & \multicolumn{1}{c}{\textbf{0.6851}}         & 0.7200          \\ \hline
			\multicolumn{2}{|c|}{}                                                        & \multicolumn{1}{c}{MoteStrain}      & \multicolumn{1}{c}{OSULeaf}                 & \multicolumn{1}{c}{Plane}           & \multicolumn{1}{c}{Prox.phal.outl.ageGroup} & Prox.phal.TW    \\ \hline
			\multicolumn{1}{|c|}{\ding{53}}                        & \checkmark                            & \multicolumn{1}{c}{0.7169}          & \multicolumn{1}{c}{0.7492}                  & \multicolumn{1}{c}{0.9264}          & \multicolumn{1}{c}{0.8003}                  & 0.8003          \\ \cline{1-2}
			\multicolumn{1}{|c|}{\checkmark}                        & \ding{53}                            & \multicolumn{1}{c}{0.7426}          & \multicolumn{1}{c}{0.7525}                  & \multicolumn{1}{c}{0.9455}          & \multicolumn{1}{c}{0.8051}                  & \textbf{0.8665} \\ \cline{1-2}
			\multicolumn{1}{|c|}{\ding{53}}                        & \ding{53}                            & \multicolumn{1}{c}{0.7393}          & \multicolumn{1}{c}{0.7546}                  & \multicolumn{1}{c}{\textbf{0.9505}} & \multicolumn{1}{c}{0.8014}                  & 0.8564          \\ \cline{1-2}
			\multicolumn{1}{|c|}{\checkmark}                        & \checkmark                            & \multicolumn{1}{c}{\textbf{0.7740}} & \multicolumn{1}{c}{\textbf{0.7546}}         & \multicolumn{1}{c}{0.9373}          & \multicolumn{1}{c}{\textbf{0.8091}}         & 0.8606          \\ \hline
	\end{tabular}}\vskip 0.1 in
\end{table*}

\subsubsection{Influence of the Cluster Distribution Losses}
\label{sec:test_distribution_loss}
In this section, we evaluate the influence of the cluster distribution losses in DTCC. Note that in the proposed DTCC method, the cluster distribution losses are enforced on both of the original view and the augmented view. As shown in Tables~\ref{table:cluster distribution nmi} and \ref{table:cluster distribution ri}, the proposed DTCC method using both cluster distribution losses outperforms the variants with zero or one cluster distribution loss on most of the datasets, which demonstrate the benefit of enforcing the cluster distribution learning on both the original and augmented views in DTCC.

\section{Conclusion}
\label{sec:conclusion}
In this paper, we propose a novel deep clustering approach termed DTCC for time series data, which integrates the temporal reconstruction via auto-encoders, the cluster distribution learning, and the dual contrastive learning into a unified framework. Particular, two parallel views (with two identical temporal auto-encoders) are formulated, where the first view corresponds to the original samples and the second view corresponds to the augmented samples. To strengthen the cluster structures in the representations learned via the auto-encoder, we incorporate the cluster structure learning with a soft $k$-means objective on both the original and augmented views. Further, the dual contrastive learning is integrated to learn more discriminative features by exploiting the sample-wise and augmentation-wise relationships as well as the global cluster-wise contrastiveness, which is able to enforce the instance-level contrastive learning and the cluster-level contrastive learning simultaneously. By modeling the auto-encoder reconstruction loss, the cluster distribution loss, and the dual contrastive losses into an overall objective, the network architecture can be trained in a self-supervised manner and the final clustering can thus be obtained for the time series data. Experiments on multiple time series datasets have confirmed the superiority of the proposed DTCC approach over the state-of-the-art time series clustering approaches.

\section*{Acknowledgments}
This work was supported by the NSFC (61976097 \& 62276277) and the Natural Science Foundation of Guangdong Province (2021A1515012203).

\bibliographystyle{elsarticle-num-names}
\bibliography{refs_2021}

\newpage

\end{document}